\documentclass[runningheads]{llncs}

\usepackage{graphicx} 
\usepackage{amsmath}  
\usepackage{amssymb}  
\usepackage{hyperref} 
\usepackage{booktabs} 
\usepackage{xcolor}   
\usepackage{subcaption}

\begin{document}

\title{Maximising the Set-Piece Return: Optimising Football Corner Tactics with Graph Reinforcement Learning}

\titlerunning{Maximising the Set-Piece Return}


\author{Sean Groom\inst{1,2} \and
Michael Groom\inst{3} \and
Francisco Belo\inst{2} \and
Axl Rice\inst{2} \and
Liam Anderson\inst{4} \and
Victor-Alexandru Darvariu\inst{3} \and
Shuo Wang\inst{1}\thanks{Corresponding author. Email: \texttt{s.wang.2@bham.ac.uk}.}}

\authorrunning{S. Groom et al.}

\institute{School of Computer Science, University of Birmingham, Birmingham, UK \and
Nottingham Forest Football Club, Nottingham, UK \and
Oxford Robotics Institute, University of Oxford, Oxford, UK \and
School of Sport, Exercise and Rehabilitation Sciences, University of Birmingham, Birmingham, UK
}

\maketitle

\begin{abstract}

Machine learning is increasingly employed for the evaluation of football tactics.
However, existing approaches focus on characterising historical actions or analyst-specified counterfactual scenarios. 
In this work, we seek to go beyond the imitation of historically observed patterns towards discovering new generalisable player configurations and strategies. 
To tackle this, we focus on optimising corner kick routines, and formulate a decision-making problem in which a central policy makes adjustments to attacking player positions and velocities to maximise first contact shot probability. 
Unlike classic optimisation that solves for isolated setups, we contribute a reinforcement learning architecture operating on graph-structured data that yields a general policy for adjusting arbitrary starting player positions. 
Evaluated on over 3,000 Premier League corners, our approach strongly outperforms baseline optimisation techniques under matched inference budgets. 
Our results suggest that graph reinforcement learning can shift set-piece analysis from historical evaluation and imitation towards reward-driven tactical discovery.

\end{abstract}

\section{Introduction}
\label{sec:intro}

The widespread adoption of machine learning in sports analytics, driven by the increasing availability of event and tracking data, has transformed tactical analysis and player evaluation~\cite{davis2024methodology}. 
In association football, modern datasets have expanded the analytical lens from traditional video analysis to encompass machine learning-based, multi-agent evaluation methods \cite{Spatio-Temporal-Analysis-of-Team-Sports,match-analysis-big-data-and-tactics-current-trends-in-elite-soccer}. Foundational models such as Expected Goals \cite{lucey2015quality,anzer2021goal}, Pitch Control \cite{RN31,spearman2018beyond} and Expected Possession Value (EPV) \cite{fernandez2021framework,Rudd2011MarkovSoccer,Singh2018xT,Cervone2016EPV,VAEP}, are immensely valuable for quantifying spatial ownership and the value of dynamic game states.

While highly effective at characterising and diagnosing historical player and coach decisions, their applicability does not extend to answering counterfactual queries about what \textit{would have happened} had a different decision been taken. Recent works have therefore adopted Markov Decision Process (MDP) formulations of football to explicitly reason about the value of alternative player actions or tactical decisions~\cite{vanroy2023markov,RahimianActionRL,nakahara2023action}. However, these approaches remain primarily tied to the historical state-action distribution. While they support counterfactual evaluation, they do not enable reward-driven \textit{discovery} of novel configurations that may never have appeared in the data. Additionally, these works study open play, which is subject to large variations in tactics and player actions. 

In contrast to open play, corner kicks provide an ideal environment for automated tactical exploration. Unlike the fluid phase shifts of open play, a corner kick is a highly structured, static restart with definitive start states. Furthermore, corners are heavily scripted, rely on repeatable tactical schemes, and account for a significant proportion of total goals (approximately 16\% in the 2025/26 English Premier League season). The current state-of-the-art in automated set-piece tactic generation, TacticAI \cite{wang2024tacticai}, leverages geometric deep learning to produce realistic, expert-preferred spatial adjustments. Yet, a \textit{critical gap} remains: such architectures optimise for reconstruction loss, conditioning outputs to imitate \textit{historically observed} player trajectories. While this imitation ensures generative realism, it inherently restricts the discovery of novel, physically plausible tactical innovations to the boundaries of the historical training distribution.

\begin{figure}[t]
    \centering
    \includegraphics[width=0.49\textwidth]{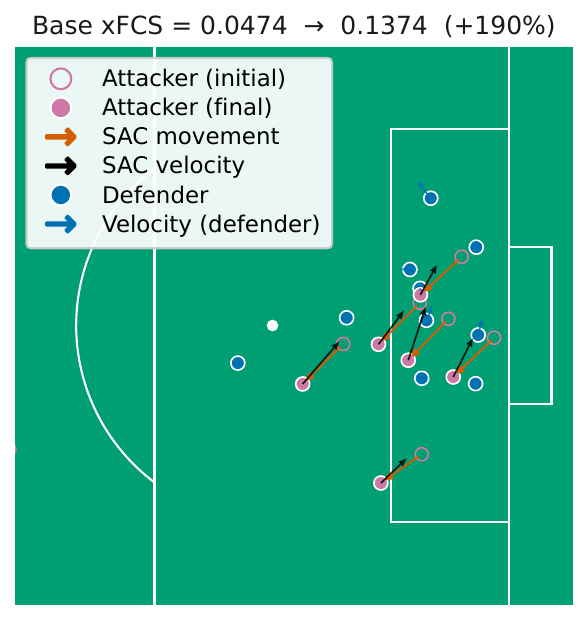}
    \hfill
    \includegraphics[width=0.49\textwidth]{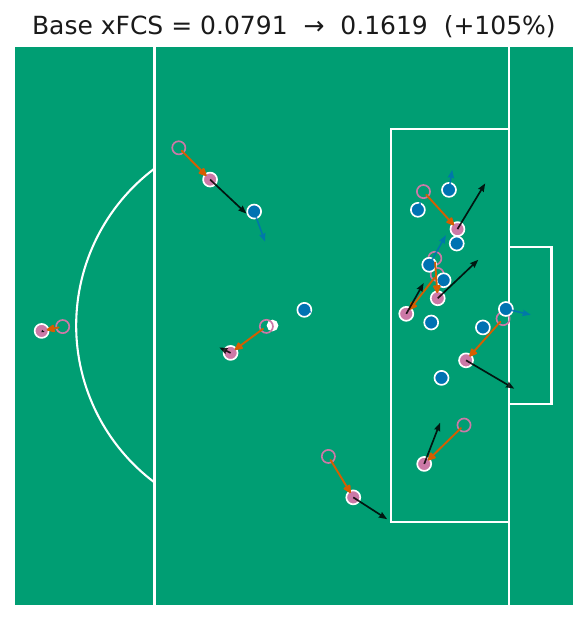}
    
    \caption{Our approach for set-piece optimisation applied to held-out Premier League test data. We take historical corner configurations as input and adjust the attacking team's starting positions (pink circles) and velocities (black arrows) to increase the Expected First Contact Shot Probability by 190\% and 105\%. These variations translate easily into actionable coaching instructions, highlighting the framework's capacity for \textit{automated tactical discovery}.}
    \label{fig:sac_adjustments}
\end{figure}

In this work, we shift the objective of automated set-piece design from the imitation of historical routines to true reward-driven optimisation. This captures the dilemma faced by coaches between \textit{exploiting} reliable, well-rehearsed routines versus \textit{exploring} novel variations that opposition defences have yet to encounter, naturally paralleling one of the main challenges in Reinforcement Learning (RL)~\cite{sutton2018reinforcement}. Our core contributions are:

\begin{enumerate}
    \item \textbf{Problem Formulation:} We formalise corner kick optimisation as an MDP. A central decision-maker is tasked with adjusting attacking player positions and velocities based on the starting state when the corner is taken. To quantify success and provide rewards, we propose an objective function (Expected First Contact Shot probability, $xFCS$), based on a predictive Graph Neural Network (GNN) model trained on Premier League data.
    \item \textbf{Computational Methodology:} We contribute a Graph Reinforcement Learning (Graph RL)~\cite{darvariu2024grl} method for learning a general policy to adjust attacking player positions in arbitrary corner states. Concretely, we integrate GNN embeddings with deep RL, evaluating both SAC \cite{haarnoja2018soft} and PPO \cite{schulman2017proximal}. 
    \item \textbf{Empirical Evaluation:} We present an evaluation on a dataset of over 3,000 Premier League corners. Results show that our method yields significantly higher-reward tactical variations compared to traditional optimisation techniques (Random Search and Simulated Annealing~\cite{kirkpatrick1983optimization}) with matched inference computational budgets. Furthermore, our approach maintains competitive performance even when the metaheuristics are granted vastly disproportionate search budgets equal to the RL agent's upfront training cost.
\end{enumerate}

\begin{figure}[t]
    \centering
    \includegraphics[trim={0cm 0cm 38cm 0cm},clip,width=0.95\columnwidth]{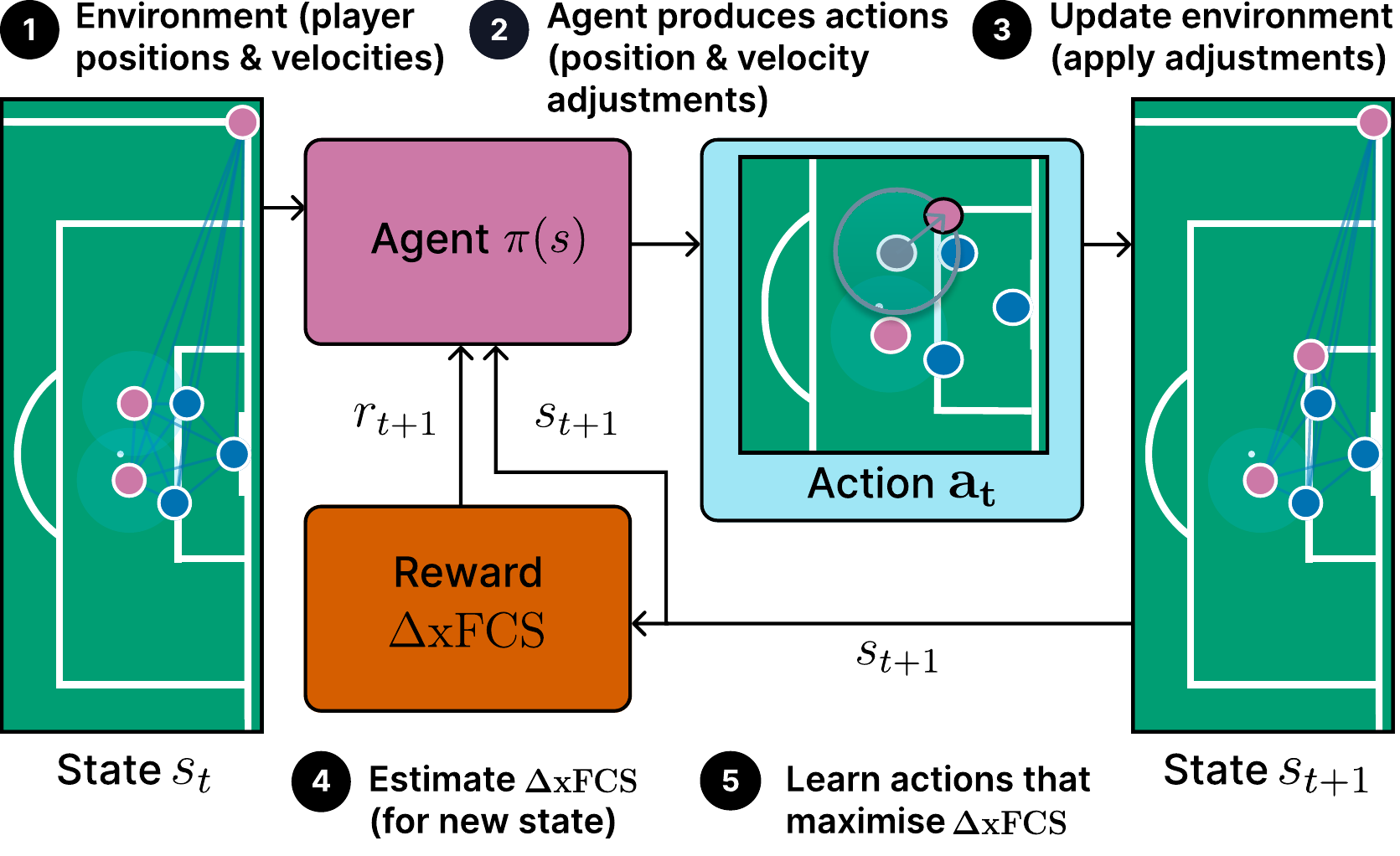}
    \caption{Illustration of one environment step underlying our set-piece optimisation approach. A step is performed for each attacking player, resulting in a final configuration that maximises $xFCS$ in expectation.} 
    \label{fig:method}
\end{figure}

\section{Methods}
\label{sec:methods}

\subsection{Problem Formulation}




We formulate the optimisation of corner kick routines as a Markov Decision Process comprising the tuple $(\mathcal{S}, \mathcal{A}, \mathcal{P}, \mathcal{R}, \gamma)$. This framing allows an RL agent to discover configurations that directly maximise expected tactical yield. 

To provide an intuitive overview, Figure \ref{fig:method} illustrates a single agent-environment interaction step. The environment initialises a historical corner and passes the graph \textit{state} $s_t \in \mathcal{S}$ to the agent. The \textit{policy} $\pi(s)$ outputs an action specifying the required positional and velocity adjustments $a_t \in \mathcal{A}$ for an attacker. The deterministic \textit{transition function} $\mathcal{P}$ updates the pitch to the new state $s_{t+1}$. The \textit{reward function} $\mathcal{R}$ calculates the objective function change against the frozen defence to return the reward $r_t$. The \textit{discount factor} $\gamma \in [0,1]$ balances immediate spatial improvements with long-term expected returns. We next formally detail the states, actions, and rewards.

\vspace{0.3em}
\noindent \textbf{States $\mathcal{S}$.} To capture the spatial dynamics and permutation invariance of the 22 players on the pitch, we define the state $s \in \mathcal{S}$ as a fully-connected homogeneous graph $\mathcal{G} = (\mathcal{V}, \mathcal{E})$. The node set $\mathcal{V}$ contains the $11$ attacking and $11$ defending players. Each node $v_i$ carries a 12-dimensional feature vector encompassing scaled spatial coordinates ($x, y$), velocity vectors ($u_x, u_y$), physical attributes (height, weight), binary role identifiers (corner taker, defending goalkeeper, delivery type, attacking team), and a binary feature indicating which player's position is to be adjusted in this state. In terms of the \textit{edges} $\mathcal{E}$, the graph is fully connected, and edges $e_{ij}$ carry a single feature consisting of a binary flag indicating whether the source and target nodes belong to the same team.

\vspace{0.3em}
\noindent \textbf{Actions $\mathcal{A}$.} Optimising the spatial configuration occurs via a sequential update process across the attacking players. At each timestep $t$, the policy outputs a continuous action vector $a_t$ that maps to two physical components: a spatial displacement $\Delta p_t = [\Delta x, \Delta y]$ and a velocity delta $\Delta u_t = [\Delta u_x, \Delta u_y]$. Within the environment, these continuous outputs are bounded to enforce realistic movement limits, restricting maximum per-step displacement to $\pm 2\text{m}$ and velocity changes to $\pm 2\text{m/s}$. Following the update step, if an attacker's resulting velocity vector magnitude $\|u_{\text{adj}}\|_2$ exceeds $5.2\text{m/s}$ (the $95^{\text{th}}$ percentile of all observed player speeds within the tracking data) the vector is proportionally scaled down to the safety boundary. This ensures that the policy cannot reach artificially high-reward states by recommending physiologically impossible sprints.


\vspace{0.3em}
\noindent \textbf{Rewards ${\mathcal{R}}$.} The reward function $\mathcal{R}(s)$ provides a dense signal quantifying the value of a given spatial configuration. We calculate this using a frozen, pre-trained GNN that maps a scene to the aforementioned $xFCS$ metric. The model decomposes the probability of a shot into two conditional factors, mirroring the causal sequence of a set piece: the delivery, the first contact, and the subsequent shot. The probability of a shot occurring given a specific scene is formulated as:
\begin{equation}
    xFCS(s) = \sum_{i \in \text{attackers}} P(\text{contact} = i \mid s) \times P(\text{shot} \mid \text{contact} = i, s).
\end{equation}
This factorisation is modelled by two separate GNNs: a reception probability model which performs corner recipient node classification, and a conditional shot model given attacker $i$ wins first contact. Extensive details regarding the architecture, training pipeline, and validation of this baseline reward model are provided in \cite{groom2026machinelearningframeworkoff}. Within the RL environment, the step reward $r_t$ is defined as the incremental change in the value of the spatial configuration following the kinematic adjustments of the attacking players:
\begin{equation}
    r_t = xFCS(s_{t}) - xFCS(s_{t-1}).
\end{equation}
This dense, delta-based formulation inherently penalises the policy for moving players into sub-optimal configurations and rewards continuous spatial optimisation towards configurations that maximise gains in $xFCS$.

\subsection{Architecture and Implementation}
\label{sec:archimpl}

\noindent \textbf{Parallelised JAX Environment}. We implement the above problem formulation as an RL environment and simulator. We opt for JAX~\cite{jax2018github} to enable significant hardware acceleration. This is particularly relevant if set-piece optimisation is to be deployed in a time-sensitive analysis or live match context, as exploring many configurations benefits from parallelised vectorisation. Our simulator is built upon a 2D physics engine~\cite{matthews2024kinetix} that supports continuous-time dynamics, which can be exploited in future work.

\vspace{0.3em}
\noindent \textbf{Deep RL Algorithms}. As RL algorithms, we consider Proximal Policy Optimization (PPO)~\cite{schulman2017proximal} and Soft Actor-Critic (SAC)~\cite{haarnoja2018soft}. The former is a widely used on-policy actor-critic algorithm with a clipped surrogate objective for training stability. The latter uses the objective $\mathcal{J}(\pi) = \mathbb{E}[\sum R(s,a) + \alpha \mathcal{H}(\pi(\cdot|s))]$ that combines expected returns and a maximum-entropy component. This forces the agent to explore a diverse array of valid tactical adjustments rather than collapsing prematurely to a single local maximum.

\vspace{0.3em}
\noindent \textbf{Graph Neural Network Representations}. The graph state $s_t$ is passed through a Graph Attention Network v2 (GATv2)~\cite{velivckovic2018graph,brody2022attentive} to capture non-local, relational player interactions. The resulting embedding $\phi(s_t)$ of the state is the input that is fed to the actor and critic networks. Specifically, the actor and critic networks each consist of two GATv2 layers with separate, unshared parameters. GATv2 layers are also used in the reward model $\mathcal{R}$. We note that the reward model for predicting shot probabilities is trained on historical data and frozen; while the parameters of the GATv2 layers for obtaining state representations $\phi(s_t)$ evolve throughout the RL training process.

\section{Experiments and Results}
\label{sec:experiments}

\subsection{Evaluation Protocol and Training Process}
To evaluate the generalisability of our Graph RL method, we utilise a dataset of 3,223 corner kicks from the English Premier League. Rather than a randomised cross-validation split, we enforce a strict 80/20 sequential temporal split. Agents are trained exclusively on the chronologically earliest 80\% corners of the season and evaluated zero-shot on the final 20\% of unseen test corners. This constraint prevents temporal data leakage and mimics the operational reality of a football season, showing how the learned policy behaves when applied to late-season structural shifts and evolutionary tactical adjustments adopted by teams.

\begin{figure}[t]
    \centering
    \includegraphics[width=0.75\textwidth]{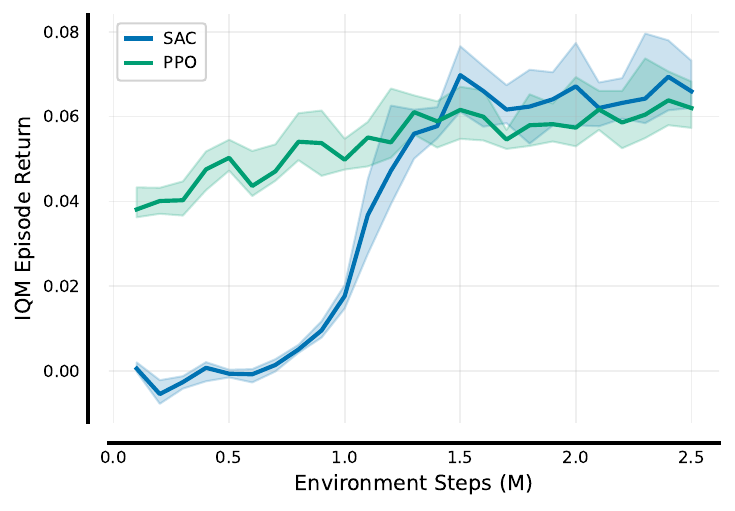}
   
    \caption{Learning curves showing performance on unseen corner kicks as a function of environment steps for the two considered RL algorithms.}
    \label{fig:training_curves}
\end{figure}

During the training phase, the RL agents interact with the JAX environment to maximise the expected $\Delta xFCS$. As illustrated in Figure \ref{fig:training_curves}, we train both the SAC and PPO agents for 2.5 million environment steps. To ensure robustness and account for training variance, we train 8 independent random seeds for each algorithm. Evaluating the policies on the held-out test set at regular intervals reveals that test-set reward plateaus near the 2.5 million step mark, with SAC converging to a peak mean improvement of approximately $+0.068\ xFCS$, slightly outperforming PPO which reaches $+0.063\ xFCS$. This convergence point informs our stopping criterion and forms the basis for our total computational budget calculations.

\subsection{Baselines and Evaluation Budgets}
We benchmark the learned RL policies against two traditional, gradient-free metaheuristic baselines:
\begin{enumerate}
    \item \textbf{Random Search (RS):} Uniformly samples continuous kinematic vectors within the defined bounding limits, acting as a brute-force spatial exploration baseline. For evaluation, we compute the mean performance across the 8 RS seeds for each individual corner; this RS mean serves as the baseline score against which all other approaches are normalised.
    \item \textbf{Simulated Annealing (SA):} Iteratively perturbs the spatial coordinates of individual attacking players in sequence, utilizing a decaying geometric temperature schedule to probabilistically accept sub-optimal adjustments and escape local minima.
\end{enumerate}

\noindent Comparing a fully trained neural network against metaheuristic search algorithms requires establishing fair computational budgets. Therefore, we evaluate the baselines under two distinct regimes. Under \textit{Inference Compute Matching}, the metaheuristics have a search budget that matches the computational cost of a single episode executed by the RL agent, reflecting an operational time-sensitive scenario. In \textit{Total Training Compute Matching}, the metaheuristics are granted an immense search budget equivalent to the \textit{entire} 2.5 million environment step cost required to train the RL agent, prorated across the test set. This serves as an approximate upper bound on the performance of the RL agent.

\begin{figure}[t]
    \centering
    \includegraphics[width=0.49\textwidth]{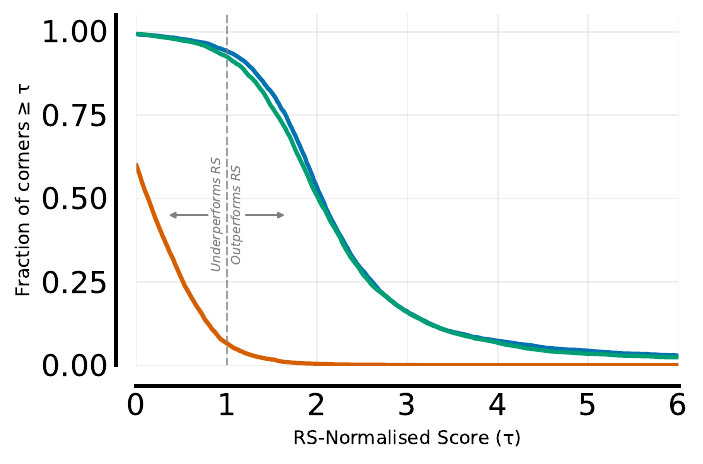}
    \hfill
    \includegraphics[width=0.49\textwidth]{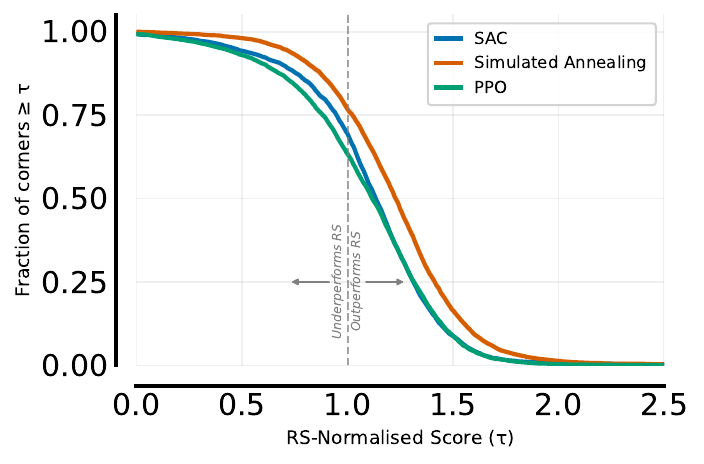}
    
    \caption{Performance across the test set in the \textit{Inference Compute Matching} (left) and \textit{Total Training Compute Matching} (right) regimes. After training, RL significantly outperforms traditional optimisation methods under the same inference time budget.}
    \label{fig:perf_profiles}
\end{figure}

\subsection{Performance Profiles}

We use the \texttt{rliable} library~\cite{agarwal2021deep} to construct performance profiles (Figure \ref{fig:perf_profiles}) that visualise the fraction of corners where a method achieves a relative performance score greater than or equal to a threshold $\tau$. To ensure a fair aggregation across the dataset, we must account for varying task difficulty; some defensive configurations natively possess more exploitable spatial flaws than others. Therefore, we define $\tau$ as the ratio of a method's $\Delta xFCS$ to the RS baseline mean for that specific corner. This standardises performance regardless of the raw reward scale: $\tau = 1.0$ indicates parity with RS, $\tau < 1.0$ indicates underperformance, and $\tau = 2.0$ indicates the algorithm generated twice the danger of the RS baseline. To ensure numerical stability and prevent $\tau$ from exploding in scenarios where RS discovers negligible improvements, we enforce a minimum denominator of $0.001$. The tight shaded regions in the plot represent 95\% stratified bootstrap confidence intervals.

Figure \ref{fig:perf_profiles}(a) evaluates the architectures under \textit{Inference Compute Matching}, i.e., realistic operational constraints where the upfront RL training cost is amortised. In a production environment, generating a tactic via a forward pass through the SAC/PPO policy is relatively computationally inexpensive. 
Under these strict latency constraints, the generalised RL policies dominate. Because the SAC and PPO agents apply a policy informed by prior training rather than searching from scratch, they outperform RS in 94.2\% and 92.6\% of evaluated scenarios, respectively. Conversely, SA collapses under the restricted budget, outperforming RS in only 6.6\% of cases. This confirms that Graph RL is uniquely equipped to provide high-quality tactical adjustments in real-time.

Figure \ref{fig:perf_profiles}(b) shifts the lens to evaluate the methods under \textit{Total Training Compute Matching}.
Under these conditions, SA treats each corner as an isolated optimisation puzzle. Given 970 reward calls per individual corner, SA acts as a powerful per-scenario optimiser, outperforming RS in 76.8\% of cases, compared to 69.4\% for the generalised SAC policy and 63.3\% for the PPO policy. While SA can outperform RL when granted a disproportionate, non-real-time compute budget, this approach scales poorly. The metaheuristic algorithm learns no overarching tactical principles and must restart its search from the beginning for every new scenario.

\subsection{Quantitative Performance Evaluation}
Beyond comparative search efficiency, we evaluate the tactical value generated across the 645 unseen test corners (two examples were visualised in Figure~\ref{fig:sac_adjustments}). The generalised RL policies demonstrate strong reliability: both SAC and PPO successfully identify advantageous kinematic adjustments ($\Delta xFCS > 0$) in over 99\% of the evaluated scenarios. In absolute terms, the SAC policy yields a mean improvement of $+0.068\ xFCS$ (median $+0.046$) per corner. Given the low-scoring nature of professional football, systematically increasing the expected number of shots from set pieces by this margin could represent a substantial competitive advantage. 
Under operational latency constraints (the inference budget), the SAC policy achieves a mean normalised score of $2.40$ (median $2.05$). This indicates that, on average, a single forward pass of the policy generates more than double the tactical value of a compute-matched RS. Even when the metaheuristic baselines are granted the disproportionate total training compute budget, SA only achieves a mean normalised score of $1.22$ against the expanded RS baseline compared to SAC and PPO scoring $1.11$ and $1.08$, respectively.



\section{Discussion and Future Work}
\label{sec:discussion}

Qualitative inspection of the generated routines reveals several consistent tactical preferences exhibited by the SAC agent. Most notably, the policy strongly favours dynamic routines. Rather than maintaining attackers in static, high-value zones (such as the centre of the six-yard box), the agent frequently displaces attackers away from their markers only to have them attack the vacated space at high velocities. This aligns with the physical reality of set pieces, where arriving in space at speed generates higher-quality shot opportunities than contested, stationary headers. 

However, evaluating the generated tactics also highlights the potential limits of a single-frame reward model. We hypothesise that while elite cooperative routines, such as deliberate ``pinning" or ``pick-and-roll" blocking schemes \cite{casal2015analysis,vaquera2016exploration,maneiro2021observational}, exist within the historical training set, they are inherently temporal events that unfold over a sequence of timesteps. Because the $xFCS$ model evaluates a single spatial snapshot composed of positions and velocities, it may struggle to definitively distinguish a coordinated blocking manoeuvre from two players simply being in close proximity. Consequently, we suspect the reward model learns a smoothed statistical expectation over these ambiguous spatial configurations. Potentially lacking a sharp, definitive signal to reward intricate blocking schemes, the RL agent may naturally gravitate toward exploiting unambiguous tactical signals that the model can confidently measure, such as generating high-velocity arrivals into undisputed space near the goal.

While metaheuristic methods such as SA are competitive in raw performance within our current formulation, this parity is largely a function of the environment's constrained dimensionality. The present optimisation task was restricted to single-team kinematic adjustments against a static defence in order to provide a proof-of-concept validation. If the environment is expanded to a reactive, multi-timestep domain, such as an adversarial Multi-Agent Reinforcement Learning (MARL) \cite{albrecht2024marl} setting where defenders dynamically adjust to counteract attacking movements, the growth of the search space would render memoryless metaheuristics computationally intractable. Conversely, our Graph RL method is better equipped to scale to these highly relational, reactive environments. Future work will move from static optimisation to adversarial MARL paradigms.

\section{Conclusion}
\label{sec:conclusion}

In this work, we aimed to transition the computational analysis of football set pieces from historical evaluation and imitation toward active, reward-driven tactical discovery. We formalised corner kick optimisation as a kinematically constrained MDP and contributed a Graph RL method integrating GNN state embeddings with continuous control policies (SAC and PPO). Our evaluation on more than 3,000 Premier League corners demonstrates that a generalised policy can effectively optimise attacking spatial configurations without relying on exhaustive, per-scenario metaheuristic search. Our approach yields significant improvements in Expected First Contact Shot Probability, strongly outperforming traditional optimisation baselines under realistic operational inference constraints. While the current methodology could provide immediate value for optimising set-piece design, future work will seek to expand the scope of tactical generation to adversarial, reactive environments.

\bibliographystyle{splncs04}

\bibliography{references} 

@inproceedings{spearman2018beyond,
  title={Beyond expected goals},
  author={Spearman, William},
  booktitle={MIT Sloan Sports Analytics Conference},
  pages={1--17},
  year={2018},
}

@article{anzer2021goal,
  title={A goal scoring probability model for shots based on synchronized positional and event data in football (soccer)},
  author={Anzer, Gabriel and Bauer, Pascal},
  journal={Frontiers in Sports and Active Living},
  volume={3},
  pages={624475},
  year={2021},
  publisher={Frontiers Media SA},
}

@article{fernandez2021framework,
  title={A framework for the fine-grained evaluation of the instantaneous expected value of soccer possessions},
  author={Fern{\'a}ndez, Javier and Bornn, Luke and Cervone, Dan},
  journal={Machine Learning},
  volume={110},
  pages={1389--1427},
  year={2021},
  publisher={Springer},
}

@article{nakahara2023action,
  title={Action valuation of on-and off-ball soccer players based on multi-agent deep reinforcement learning},
  author={Nakahara, Hitoshi and Tsutsui, Kazuya and Takeda, Kazuya and Fujii, Keisuke},
  journal={IEEE Access},
  volume={11},
  pages={131237--131244},
  year={2023},
  publisher={IEEE},
}

@article{wang2024tacticai,
  title={{TacticAI}: an {AI} assistant for football tactics},
  author={Wang, Zhe and Veli{\v{c}}kovi{\'c}, Petar and Hennes, Daniel and others},
  journal={Nature Communications},
  volume={15},
  number={1},
  pages={1906},
  year={2024},
}

@inproceedings{velivckovic2018graph,
  title={Graph attention networks},
  author={Veli{\v{c}}kovi{\'c}, Petar and Cucurull, Guillem and Casanova, Arantxa and Romero, Adriana and Liò, Pietro and Bengio, Yoshua},
  booktitle = {International Conference on Learning Representations (ICLR)},
  year={2018}
}

@inproceedings{brody2022attentive,
  title={How attentive are graph attention networks?},
  author={Brody, Shaked and Alon, Uri and Yahav, Eran},
  booktitle = {International Conference on Learning Representations (ICLR)},
  year={2022}
}

@inproceedings{haarnoja2018soft,
  title={Soft actor-critic: Off-policy maximum entropy deep reinforcement learning with a stochastic actor},
  author={Haarnoja, Tuomas and Zhou, Aurick and Abbeel, Pieter and Levine, Sergey},
  booktitle={International Conference on Machine Learning (ICML)},
  year={2018},
}

@article{schulman2017proximal,
  title={Proximal policy optimization algorithms},
  author={Schulman, John and Wolski, Filip and Dhariwal, Prafulla and Radford, Alec and Klimov, Oleg},
  journal={arXiv preprint arXiv:1707.06347},
  year={2017}
}

@software{jax2018github,
  author = {James Bradbury and Roy Frostig and Peter Hawkins and Matthew James Johnson and Yash Katariya and Chris Leary and Dougal Maclaurin and George Necula and Adam Paszke and Jake Vander{P}las and Skye Wanderman-{M}ilne and Qiao Zhang},
  title = {{JAX}: composable transformations of {P}ython+{N}um{P}y programs},
  url = {http://github.com/jax-ml/jax},
  version = {0.3.13},
  year = {2018},
}

@article{Spatio-Temporal-Analysis-of-Team-Sports,
   author = {Gudmundsson, J. and Horton, M.},
   title = {Spatio-Temporal Analysis of Team Sports},
   journal = {{ACM} Computing Surveys},
   volume = {50},
   number = {2},
   year = {2017},
}

@article{match-analysis-big-data-and-tactics-current-trends-in-elite-soccer,
   author = {Memmert, D. and Rein, R.},
   title = {Match Analysis, Big Data and Tactics: Current Trends in Elite Soccer},
   journal = {Deutsche Zeitschrift für Sportmedizin},
   volume = {69},
   number = {3},
   pages = {65-72},
   year = {2018},
}

@inproceedings{lucey2015quality,
  author    = {Lucey, Patrick and Bialkowski, Alina and Monfort, Mathew and Carr, Peter and Matthews, Iain},
  title     = {{``Quality vs Quantity"}: Improved Shot Prediction in Soccer using Strategic Features from Spatiotemporal Data},
  booktitle = {MIT Sloan Sports Analytics Conference},
  year      = {2015},
}

@inproceedings{RN31,
   author = {Taki, T. and Hasegawa, J.},
   title = {Visualization of dominant region in team games and its application to teamwork analysis},
   booktitle = {International Conference of the Computer Graphics Society (CGI)},
   pages = {227-235},
   year = {2000},
}

@inproceedings{Rudd2011MarkovSoccer,
  author    = {Sarah Rudd},
  title     = {A Framework for Tactical Analysis and Individual Offensive Production Assessment in Soccer Using Markov Chains},
  booktitle = {New England Symposium on Statistics in Sports (NESSIS)},
  year      = {2011},
  note      = {Conference presentation; slides.},
  url       = {https://nessis.org/nessis11/rudd.pdf},
  urldate   = {2025-09-16}
}

@misc{Singh2018xT,
  author       = {Karun Singh},
  title        = {Introducing Expected Threat {(xT)}},
  year         = {2018},
  howpublished = {\url{https://karun.in/blog/expected-threat.html}},
  urldate      = {2025-09-16}
}

@article{Cervone2016EPV,
  author  = {Cervone, Daniel and D'Amour, Alex and Bornn, Luke and Goldsberry, Kirk},
  title   = {A Multiresolution Stochastic Process Model for Predicting Basketball Possession Outcomes},
  journal = {Journal of the American Statistical Association},
  year    = {2016},
  volume  = {111},
  number  = {514},
  pages   = {585--599},
}

@inproceedings{VAEP,
author = {Decroos, Tom and Bransen, Lotte and Van Haaren, Jan and Davis, Jesse},
year = {2019},
pages = {1851-1861},
title = {Actions Speak Louder than Goals: Valuing Player Actions in Soccer},
booktitle={ACM SIGKDD International Conference on Knowledge Discovery \& Data Mining (KDD)}
}

@article{davis2024methodology,
  title={Methodology and evaluation in sports analytics: challenges, approaches, and lessons learned},
  author={Davis, Jesse and Bransen, Lotte and Devos, Laurens and Jaspers, Arne and Meert, Wannes and Robberechts, Pieter and Van Haaren, Jan and Van Roy, Maaike},
  journal={Machine Learning},
  volume={113},
  number={9},
  pages={6977--7010},
  year={2024},
  publisher={Springer}
}

@book{sutton2018reinforcement,
  title={Reinforcement learning: An introduction},
  author={Sutton, Richard S and Barto, Andrew G.},
  year={2018},
  publisher={MIT Press},
  address={Cambridge, MA}
}

@article{kirkpatrick1983optimization,
  title={Optimization by simulated annealing},
  author={Kirkpatrick, Scott and Gelatt Jr., C. Daniel and Vecchi, Mario P.},
  journal={Science},
  volume={220},
  number={4598},
  pages={671--680},
  year={1983},
  publisher={American Association for the Advancement of Science},
}

@article{RahimianActionRL,
   author = {Rahimian, Pegah and Toka, Laszlo},
   title = {A data-driven approach to assist offensive and defensive players in optimal decision making},
   journal = {International Journal of Sports Science and Coaching},
  volume={19},
  number={1},
  pages={245--256}
}

@article{groom2026machinelearningframeworkoff,
      title={A Machine Learning Framework for Off Ball Defensive Role and Performance Evaluation in Football}, 
      author={Sean Groom and Shuo Wang and Francisco Belo and Axl Rice and Liam Anderson},
      year={2026},
      journal={arXiv preprint arXiv:2601.00748},
}

@inproceedings{matthews2024kinetix,
      title={Kinetix: Investigating the Training of General Agents through Open-Ended Physics-Based Control Tasks}, 
      author={Michael Matthews and Michael Beukman and Chris Lu and Jakob Foerster},
      booktitle={International Conference on Learning Representations (ICLR)},
      year={2025},
}

@article{vanroy2023markov,
  title={A Markov framework for learning and reasoning about strategies in professional soccer},
  author={Van Roy, Maaike and Robberechts, Pieter and Yang, Wen-Chi and De Raedt, Luc and Davis, Jesse},
  journal={Journal of Artificial Intelligence Research},
  volume={77},
  pages={517--562},
  year={2023}
}

@article{agarwal2021deep,
  title={Deep reinforcement learning at the edge of the statistical precipice},
  author={Agarwal, Rishabh and Schwarzer, Max and Castro, Pablo Samuel and Courville, Aaron C. and Bellemare, Marc},
  journal={Advances in neural information processing systems},
  volume={34},
  pages={29304--29320},
  year={2021}
}

@article{darvariu2024grl,
  title={Graph Reinforcement Learning for Combinatorial Optimization: A Survey and Unifying Perspective},
  author={Victor-Alexandru Darvariu and Stephen Hailes and Mirco Musolesi},
  journal={Transactions on Machine Learning Research},
  year={2024},
}

@article{casal2015analysis,
  title={Analysis of corner kick success in elite football},
  author={Casal, C. A. and Maneiro, R. and Ard{\'a}, T. and Losada, J. L. and Rial, A.},
  journal={International Journal of Performance Analysis in Sport},
  volume={15},
  number={2},
  pages={430--451},
  year={2015},
  publisher={Taylor \& Francis}
}

@article{vaquera2016exploration,
  title={An exploration of ball screen effectiveness on elite basketball teams},
  author={Vaquera, A. and Garc{\'\i}a-Tormo, J. V. and G{\'o}mez Ruano, M. A. and Morante, J. C.},
  journal={International Journal of Performance Analysis in Sport},
  volume={16},
  number={2},
  pages={475--485},
  year={2016},
  publisher={Taylor \& Francis}
}

@article{maneiro2021observational,
  title={Observational Analysis of Corner Kicks in High-Level Football: A Mixed Methods Study},
  author={Maneiro, R. and Losada, J. L. and Portell, M. and Ard{\'a}, A.},
  journal={Sustainability},
  volume={13},
  number={14},
  pages={7562},
  year={2021},
  publisher={MDPI}
}

@book{albrecht2024marl,
  title={Multi-Agent Reinforcement Learning: Foundations and Modern Approaches},
  author={Albrecht, Stefano V. and Christianos, Filippos and Sch\"{a}fer, Lukas},
  year={2024},
  publisher={MIT Press},
  address={Cambridge, MA},
  isbn={978-0-262-38050-8}
}

\end{document}